\documentclass[accepted]{uai2022} 

\usepackage[american]{babel}

\usepackage{natbib} 
\usepackage{mathtools} 
\usepackage{booktabs} 
\usepackage{tikz} 
\usepackage{amsfonts}
\usepackage{csquotes}
\usepackage{caption}
\usepackage{subcaption}
\usepackage[ruled,linesnumbered]{algorithm2e}

\DeclareMathOperator*{\argmax}{arg\,max}

\newcommand{\RSA}{\mathbb{R}^{|\mathcal{S}|\times|\mathcal{A}|}}

\newcommand{\RS}{\mathbb{R}^{|\mathcal{S}|}}

\newcommand{\AdaBracket}[1]{\left(#1\right)}
\newcommand{\AdaRectBracket}[1]{\left[#1\right]}

\newcommand{\AdaAngleProduct}[2]{\left\langle#1, #2\right\rangle}

\newcommand{\sumJK}[1]{\sum_{j=#1}^{k}}

\newcommand{\KL}{D_{\!K\!L}\!\AdaBracket{\pi || \bar{\pi}}}
\newcommand{\KLindex}[2]{D_{\!K\!L}\!\AdaBracket{\pi_{#1}|| \pi_{#2}}}

\newcommand{\entropy}{\mathcal{H}\left( \pi \right)}

\newcommand{\entropyIndex}[1]{\mathcal{H}\left( \pi_{#1} \right)}

\newcommand{\greedy}[1]{\argmax_{\pi}\!\AdaAngleProduct{\pi}{#1}}
\newcommand{\greedyH}[1]{\argmax_{\pi}\!\AdaAngleProduct{\pi}{#1 - \tau\ln\pi}}

\newcommand{\BellmanH}[2]{r + \gamma P\!\AdaAngleProduct{\pi_{#1}}{Q_{#2} - \tau\ln\pi_{#1}}}

\newcommand{\Bellman}[2]{r + \gamma P\!\AdaAngleProduct{\pi_{#1}}{Q_{#2}}}

\newcommand{\eq}[1]{Eq.\,(#1)}

\newcommand{\transition}{\gamma P_{\pi}}

\newcommand{\qlog}{$q$-logarithm }

\newcommand{\qstarexp}{$q^*$-exponential }
\newcommand{\tsallis}[1]{S_q(#1)}
\newcommand{\sparse}[1]{S_2(#1)}

\newcommand{\qKLindex}[2]{D^{q}_{\!K\!L}\!\AdaBracket{\pi_{#1}|| \pi_{#2}}}
\newcommand{\qKLany}[2]{D^{q}_{\!K\!L}\!\left(#1 \left| \! \right| #2 \right)}
\newcommand{\logqstar}[1]{\ln_{q^{*}}\!#1}
\newcommand{\expqstar}[1]{\exp_{q^{*}}\!#1}

\title{$q$-Munchausen Reinforcement Learning}

\author[1]{\href{mailto:<lingwei.andrew.zhu@gmail.com>?Subject=Your paper}{Lingwei~Zhu}{}}
\author[2]{Zheng Chen}
\author[3]{Eiji~Uchibe}
\author[1]{Takamitsu~Matsubara}
\affil[1]{%
    Nara Institute of Science and Technology, Japan
}
\affil[2]{%
    Osaka University, Japan
}
\affil[3]{%
    Advanced Telecommunication Research, Japan
  }
  
\begin{document}
\maketitle

\begin{abstract}

The recently successful Munchausen Reinforcement Learning (M-RL) features implicit Kullback-Leibler (KL) regularization by augmenting the reward function with logarithm of the current stochastic policy.
Though significant improvement has been shown with the Boltzmann softmax policy, when the Tsallis sparsemax policy is considered, the augmentation leads to a flat learning curve for almost every problem considered. 
We show that it is due to the mismatch between the conventional logarithm and the non-logarithmic (generalized) nature of Tsallis entropy.
Drawing inspiration from the Tsallis statistics literature, we propose to correct the mismatch of M-RL with the help of $q$-logarithm/exponential functions.
The proposed formulation leads to implicit Tsallis KL regularization under the maximum Tsallis entropy framework.
We show such formulation of M-RL again achieves superior performance on benchmark problems and sheds light on more general M-RL with various entropic indices $q$.

\end{abstract}
  
  \section{Introduction}
  \label{sec:introduction}

Temporal difference (TD) learning nowadays encompasses a large number of important reinforcement learning (RL) algorithms that attempt to learn the optimal value function by bootstrapping the current estimate \citep{Sutton-RL2018}.
Recently proposed modifications on the TD loss often introduce additional information metrics of policies, e.g. Shannon entropy of the current policy for encouraging better exploration of the state-action space \citep{ziebart2010-phd}; or Kullback-Leibler (KL) divergence for penalizing deviation from some given baseline policy \citep{azar2012dynamic,Fox2016}.
Such augmentation trades off optimality bias for benefits such as multimodal optimal policy (Shannon entropy) \citep{haarnoja-SAC2017a} or average of past values (KL) \citep{vieillard2020leverage}.

To avoid computing the recursively defined KL regularized optimal policy that could incur large error, Munchausen RL (M-RL) \citep{vieillard2020munchausen} proposes to implicitly perform KL regularization by adding logarithm of the current stochastic policy of any kind to the TD loss.
Empirically, significant improvement upon Deep Q-Network \citep{mnih2015human} has been shown by adding the log-softmax-policy, derived as the result of Shannon entropy augmentation.
M-RL is an exemplar method of implicit KL regularization: in this setting a new action value function is defined and iterated upon, and the greedy policy with respect to which resembles the true explicitly regularized one \citep{kozunoCVI}.

Motivated by the claim that arbitrary stochastic policy could be used for the Munchausen log-policy augmentation, in many scenarios it is desired to exploit other stochastic but more \emph{compact} policies assigning probabilities only to a subset of actions.
Such policies are more well-suited to problems that require safety or parsimony on action selection: e.g. some actions under certain circumstances must have strictly zero probabilities of being selected.
This formulation naturally invites Tsallis sparse entropy \citep{Lee2018-TsallisRAL,chow18-Tsallis}.
However, when Tsallis sparse policy is used in place of the softmax in the log-policy augmentation, confusing flat learning curves emerge for almost every problem considered, contradicting the general claim.
In this paper, we view the problem as a result of the mismatch between the conventional logarithm and the non-logarithmic (generalized) nature of Tsallis entropy \citep{TsallisEntropy}: the behavior of Tsallis entropy is better described by the deformed $q$-logarithm/exponential functions rather than their standard counterparts \citep{Yamano2004-properties-qlogexp}.

To correct the mismatch, we propose to use the $q$-log-policy drawing inspiration from the Tsallis statistics literature.
Maximum Tsallis entropy framework \citep{Lee2020-generalTsallisRSS} is employed under which we show the Munchausen augmentation leads to implicit Tsallis KL regularization \citep{Furuichi2004-fundamentals-qKL}
by the help of $q$-logarithm and its $2-q$ duality \citep{Naudts2002DeformedLogarithm}.
The corrected formulation ensures the consistency between the non-logarithmic Tsallis policies and the Munchausen augmentation, again leading to improved performance as the original M-RL did.

The contribution of this paper is mainly theoretical in introducing the Tsallis KL regularization to bridge the gap between the maximum Tsallis entropy framework and implicit KL regularization put forward by M-RL.
Related work and background are provided in Section \ref{sec:related_work} and \ref{sec:preliminary}, respectively.
Section \ref{sec:proposal} introduces the proposed method to achieving implicit Tsallis KL regularization.
We validate the proposed method by several proof-of-concept experiments on benchmark problems in Section \ref{sec:experiment}, and discuss possible future directions in Section \ref{sec:conclusion}.
  
\section{Related Work}\label{sec:related_work}

\textbf{Entropy Regularization. }
Recent successful RL algorithms often feature the use of entropy regularization, which refers to augmenting the reward function with the Shannon entropy of policies as bonus \citep{ziebart2010-phd,haarnoja-SAC2017a,haarnoja-SAC2018}, or penalizing deviating from some baseline policy (which is often the previous policy) via KL divergence \citep{azar2012dynamic,Fox2016}.
It is known that the Shannon entropy augmentation renders the optimal policy stochastic and smoothens the optimization landscape \citep{ahmed19-entropy-policyOptimization}, and KL regularization groups the error by summation, which could lead to asymptotic cancellation of errors if mild assumptions such as the sequence of errors is martingale difference \citep{azar2012dynamic,vieillard2020leverage}.
This result testified to prior work that empirically averaged action values and showed to substantially improve upon the vanilla Deep Q-Network (DQN) \citep{Anschel2017-AverageDQN,Lan2020-maxminQ}.

However, a KL regularized policy recursively depends on the previous policies which poses a challenge for implementation \citep{geist19-regularized}.
Exactly computing such recursively defined policy is intractable in practice since prohibitive memory complexity will be incurred by the need of storing all previous policies.
On other hand, na\"{i}vely evaluating only one previous policy as a surrogate for the recursive dependence typically results in large greedy step errors and underperformance \citep{Nachum2017-bridgeGap,vieillard2020munchausen}.
Recently, there is a trend on implicitly performing KL regularization so its properties can be enjoyed without needing to care about its recursive nature \citep{vieillard2020munchausen,Vieillard2020Momentum}, in which Munchausen DQN (MDQN) was one of the state-of-the-art method, achieving comparable performance with distributional methods.
However, different with the claim that any stochastic policy could be used, MDQN with Tsallis entropy induced policies showed flat learning curves for even very simple control problems.
In this paper, we show that it is because of the mismatch between the logarithm and the non-logarithmic Tsallis entropy.

\textbf{Tsallis Statistics. }
In the field of statistical mechanics, it is well known that a wide range of phenomena in nature such as chaos or fractals can be successfully described by power-law theories rather than the traditional Gibbs-Boltzmann statistics \citep{Suyari2005-LawErrorTsallis}.
Tsallis entropy has hence been proposed \citep{TsallisEntropy} and successfully applied to various problems of which power law was of concern.
Tsallis entropy can be seen as a generalization of Shannon entropy, described by deformed logarithm and exponential functions \citep{Naudts2002DeformedLogarithm,Yamano2004-properties-qlogexp}.

In RL, the introduction of Tsallis entropy was quite recent, benefited greatly from the sparsity-inducing Tsallis sparse policy \citep{Martins16-sparsemax,Lee2018-TsallisRAL,chow18-Tsallis}.
It turns out that the Tsallis sparse entropy and Shannon entropy are both special cases of the maximum general Tsallis entropy framework considered in \citep{Lee2020-generalTsallisRSS}.
However, maximum Tsallis entropy methods (excluding Shannon entropy) often suffer from various errors: the closed-form sparsemax policy is inherently more sensitive to errors than softmax \citep{Lee2020-generalTsallisRSS}, while other entropies do not possess closed-form policy and entail approximation which could incur significant greedy step error.
Motivated by the error cancellation effect of KL, we introduce Tsallis KL (relative entropy) regularization into the maximum Tsallis entropy framework
 \citep{Furuichi2004-fundamentals-qKL,Ohara2007-geometryDisrtibutionsTsallis} which has not seen published results to the best of the authors' knowledge.

\section{Background}\label{sec:preliminary}

\subsection{Reinforcement Learning}\label{sec:rl}

We focus on discrete-time discounted Markov Decision Processes (MDPs) which are expressed by the quintuple $(\mathcal{S}, \mathcal{A}, P, r,  \gamma)$, where $\mathcal{S}$ and $\mathcal{A}$ denote state space and finite action space, respectively. 
$P(\cdot |s,a)$ denotes transition probability over the state space given state-action pair $(s,a)$,  
and $r(s,a) \in [0, r_{max}]$ defines the reward associated with that transition. 
When time step $t$ is of concern, we write $r_t := r(s_t, a_t)$.
$\gamma \in (0,1)$ is the discount factor.
A policy $\pi(\cdot|s)$ is a mapping from the state space to distributions over actions.
For control purposes, we define the state-action value function following policy $\pi$ as $Q_{\pi} \in \RSA$:
\begin{align}
    Q_{\pi}(s,a) = \mathbb{E}\AdaRectBracket{\sum_{t=0}^{\infty} r_t | s_0=s, a_0 = a } ,
\end{align}
where the expectation is with respect to the policy $\pi$ and transition probability.

For notational convenience, we define the inner product for any two functions $F_1,F_2\in\RSA$  as $\AdaAngleProduct{F_1}{F_2} \in \RS$. 
$\boldsymbol{1}$ denotes an all-one vector whose dimension should be clear from the context.
With the above notations,  the Bellman operator acting upon any function $Q \!\in\! \RSA$ can be defined as: $T_{\pi}Q := r + \transition Q$, where $P_{\pi}Q \!\in\! \RSA \!:=\! \mathbb{E}_{s'\sim P(\cdot|s,a),a'\sim\pi(\cdot|s')}\!\AdaRectBracket{Q(s',a')} \!=\! P\!\AdaAngleProduct{\pi}{Q}$, and the definition should be understood as component-wise.
Repeatedly applying the Bellman operator renders $Q$ converges to the unique fixed point $Q_{\pi}$, which also includes the optimal action value function $Q_*:=Q_{\pi^*}$ \citep{Bertsekas:1996:NP:560669}.
In this paper we focus on value iteration methods that perform the following loop to obtain the optimal policy:
\begin{align}
    \begin{cases}
        \pi_{k+1} = \greedy{Q_{k}}, & \\
        Q_{k+1} =   \Bellman{k+1}{k}. &
    \end{cases}
    \label{eq:PI}
\end{align}
In practice, one often parametrizes $Q$ functions by weight vectors $\theta$ and optimizes the TD loss $Q_{\theta} - r - \gamma \AdaAngleProduct{\pi}{Q_{\bar{\theta}}}$ to obtain the new action value, where $\bar{\theta}$ denotes a fixed target of $\theta$ \citep{mnih2015human}.

\subsection{Entropy Regularization}\label{sec:entropy}

Recent RL algorithms often feature the trade-off between faster convergence and optimality bias introduced by information metrics.
Specifically, Shannon entropy, KL divergence and Tsallis entropy are among the most employed, respectively defined as  $\entropy \!:=\! \AdaAngleProduct{-\pi}{\ln\pi}$,  $\KL \!:=\! \AdaAngleProduct{\pi}{\ln\pi - \ln\bar{\pi}}$ and $S_q(\pi)\!:=\! \frac{k}{q-1}\AdaBracket{ 1  -  \AdaAngleProduct{\boldsymbol{1}}{\pi^q} }$.
Note that $S_q(\pi)$ converges to $\entropy$ when $k\!=\!\frac{1}{2}, q\rightarrow 1$.
In this paper we consider the case $k \!=\!\frac{1}{2}, q\!=\!2$ for $S_q(\pi)$ which corresponds to the Tsallis sparse entropy $S_2(\pi) \!:=\! \frac{1}{2} \! \AdaAngleProduct{\pi}{(1 - \pi)}$ \citep{Lee2018-TsallisRAL,chow18-Tsallis}.
Let us associate a coefficient $\tau$ to $\entropy$, 
we can then succinctly write the soft value iteration as:
\begin{align}
  \begin{cases}
      \pi_{k+1} = \greedyH{Q_{k}}, & \\
      Q_{k+1} =   \BellmanH{k+1}{k}. &
  \end{cases}
  \label{eq:regularizedPI}
\end{align}
It is well known that $\pi_{k+1} = \exp\AdaBracket{\tau^{-1}(Q_k-V_k)}$, where $V_k \!=\! \tau\log\AdaAngleProduct{\boldsymbol{1}}{\tau^{-1}Q_k}$ is the soft value function \citep{haarnoja-SAC2017a}.
If we replace $\entropy$ with $S_2(\pi)$ in \eq{\ref{eq:regularizedPI}}, the greedy policy becomes a \emph{sparsemax} policy: $\pi_{k+1} \!=\! \AdaRectBracket{\frac{Q_k}{\tau} - \psi\AdaBracket{ \frac{Q_k}{\tau}} }_{+}$, where $[\cdot]_{+} = \max\{\cdot, 0\}$, and $\psi$ is the normalization ensuring the policy sums to 1:
\begin{align}
  \begin{split}
      \psi\AdaBracket{\frac{Q(s,\cdot)}{\tau}}  = \frac{\sum_{a\in S(s)} \frac{Q(s,a)}{\tau} - 1 }{|S{(s)}|},
  \end{split}
  \label{eq:sparse_normalization}
\end{align}
where $S(s)$ is the set of actions satisfying $1 \!+\! i\frac{Q(s,a_{(i)})}{\tau} \!>\! \sum_{j=1}^{i}\frac{Q(s,a_{(j)})}{\tau}$, $a_{(j)}$ indicates the action with $j$th largest action value, $|S(s)|$ denotes the cardinality of $S(s)$.
It is hence clear that Tsallis entropy augmentation renders policies sparse by truncating actions $a'$ that have lower values than the normalization $\frac{Q(s, a')}{\tau} < \psi\AdaBracket{\frac{Q(s, a')}{\tau}}$.
This truncation is shown in the bottom of Figure \ref{fig:qkl_examples}.
In the rest of the paper, we shall respectively call the Boltzmann softmax and Tsallis sparsemax policy as softmax and sparsemax policy in short.
We also term MDQN with sparsemax augmentation as log-sparsemax MDQN.

\subsection{Munchausen RL}\label{sec:Munchausen}

The concept of M-RL can be best described by that the authors advocate for \enquote{optimizing for the immediate reward augmented by the scaled log-policy of the agent when using any TD scheme} \citep[p.1]{vieillard2020munchausen}.
As a concrete instantiation, M-RL is implemented via Munchausen Deep Q-Netwrok (MDQN):
\begin{align}
    \begin{split}
        \begin{cases}
            \pi_{k+1} = \argmax_{\pi} \AdaAngleProduct{\pi}{Q_{k}} + \tau\entropy  \\
            Q_{k+1} = r + {\color{red}{\alpha\tau \ln\pi_{k+1} }} + \gamma P \AdaAngleProduct{\pi_{k+1}}{Q_k - {\color{blue}\tau\ln\pi_{k+1} } } , 
        \end{cases} 
    \end{split}
    \label{eq:mdqn_recursion}
\end{align}
where the blue term comes from that M-RL requires a stochastic policy and hence Shannon entropy augmentation is employed.
 The red term is the Munchausen log-policy term, with $\alpha$ an additional coefficient.

The evaluation step of \eq{\ref{eq:mdqn_recursion}} has the following equivalence:
\begin{align}
  \begin{split}
    &Q_{k+1} \!-\! {\color{black}{\alpha\tau \ln\pi_{k+1} }} = \\
    & r \!+\!   \gamma P\! \AdaAngleProduct{\!\pi_{k+1}}{Q_k \!-\! {\color{black} \alpha\tau\ln\frac{\pi_{k+1}}{\pi_k} } \!-\! \alpha\tau\ln\pi_{k} \!-\! {\color{black} (1\!-\!\alpha)\tau\ln\pi_{k+1} }  \!\!}\\
    & = r + \gamma P \big(\!\AdaAngleProduct{\pi_{k+1}}{Q_k - \alpha\tau\ln\pi_{k}} -  \\
    & \qquad \qquad \qquad \quad \alpha\tau\KLindex{k+1}{k} + (1-\alpha)\tau\entropyIndex{k+1} \!\big).
  \end{split}
  \label{eq:munchausen_derivation}
\end{align}
By defining a new action value function $Q'_{k+1}:=Q_{k+1} \!-\! \alpha\tau\ln\pi_{k+1}$, we see that the augmentation of $\alpha\tau\ln\pi_{k+1}$ led to implicit KL regularization.

\section{Tsallis Munchausen RL}\label{sec:proposal}

\subsection{Logarithm Sparsemax Policy}

\begin{figure}
  \centering
  \includegraphics[width=0.8\linewidth]{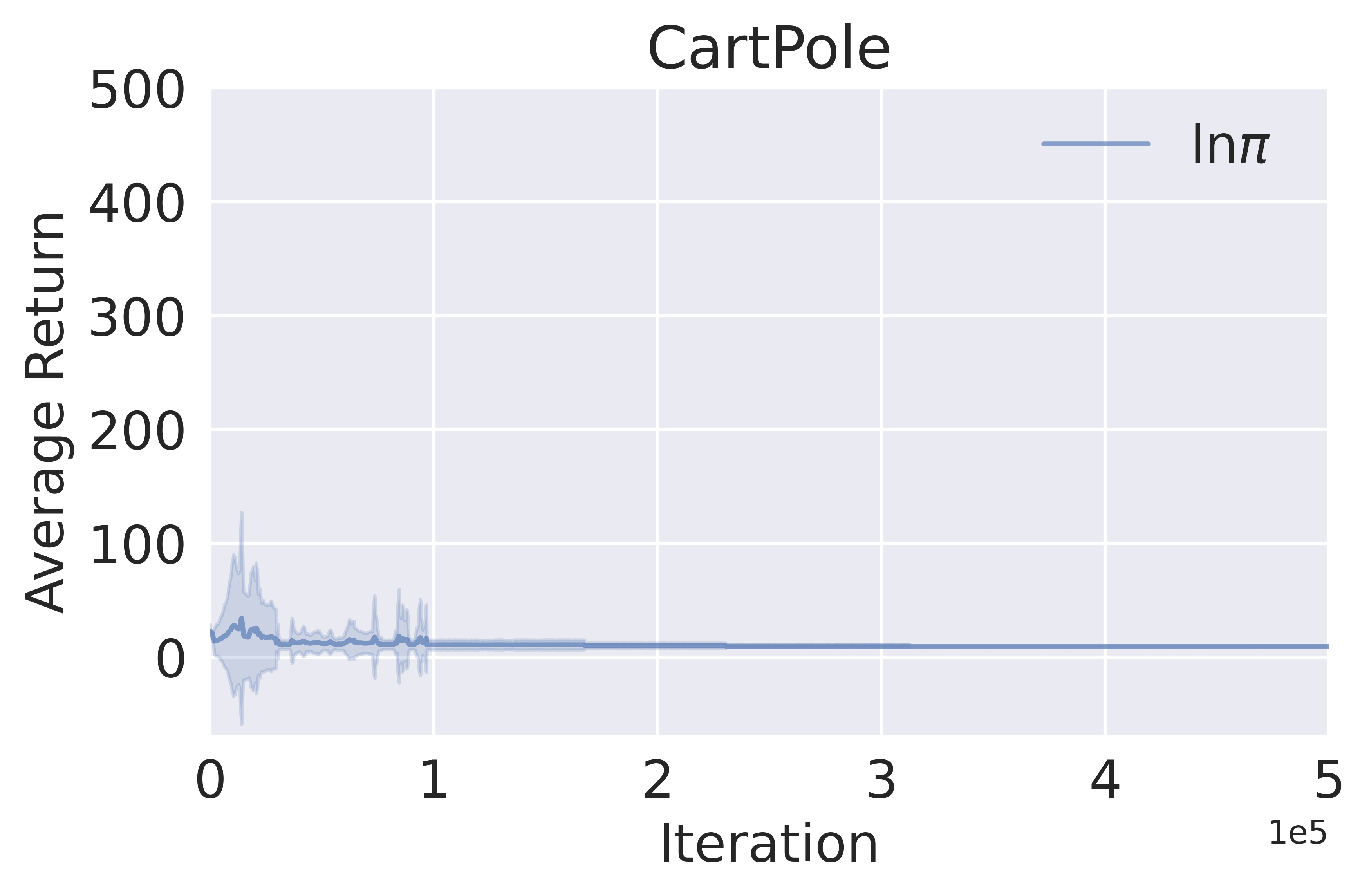}
\caption{MDQN with the standard $\ln\pi$ augmentation,
with $\pi$ being the Tsallis sparsemax policy induced by Tsallis sparse policy $S_2(\pi)$.
Mean and $\pm 1$ standard deviation are averaged over 50 independent trials.
}
\label{fig:MT_fail}
\end{figure}

While MDQN was shown to perform well with Shannon entropy, it exhibits an almost flat learning curve for even simple benchmark problems such as \texttt{CartPole-v1} in the classic control library \citep{brockman2016openai}, as can be seen from Figure \ref{fig:MT_fail}. 

The failure of MDQN in solving the task can be interpreted from two viewpoints: advantage and KL regularization.
In the advantage perspective, one may consider the Boltzmann policy \emph{encapsulates} information of the action gap \citep{Farahmand2011-actionGap} since $\pi_{k+1} = \exp\AdaBracket{\tau^{-1}(Q_k-V_k)}$, and the logarithm can hence be regarded as \emph{decapsulating} it.
When the Tsallis policy is adopted, the logarithm fails to decipher the information contained in the power-law Tsallis entropy.

The KL regularization perspective is intuitive in that the equivalence in \eq{\ref{eq:munchausen_derivation}} no longer holds with Tsallis entropy augmentation $\tau\sparse{\pi_{k+1}} \!=\! -\tau\AdaAngleProduct{\pi_{k+1}}{(1 - \pi_{k+1})}$ ($\tau$ absorbs the constant $\frac{1}{2}$).
Since no logarithm term appears, it is clear that implicit KL regularization can no longer be performed.
It is hence desired that the Tsallis entropy be written in logarithms to match the derivation in \eq{\ref{eq:munchausen_derivation}}:
\begin{align}
  Q_{k+1} = r + \alpha\tau \ln_{q}\pi_{k+1}  + \gamma P \AdaAngleProduct{\pi_{k+1}}{Q_k - \tau\ln_q\pi_{k+1}} ,
  \label{eq:mdqn_qlog}
\end{align}
where $\ln_q\pi_{k+1}$ is the assumed logarithm function capable of representing $(1-\pi_{k+1})$.
There indeed exists such logarithm functions by drawing inspiration from the Tsallis statistics literature, by which we can describe the general Tsallis entropy $\tsallis{\pi}$ via the deformed $q$-logarithm. 
We introduce it in the next sub-section.

\subsection{$q$-logarithm}\label{sec:qlog}

Tsallis entropy $S_q(\pi)$ has been heavily exploited in statistical physics to describe phenomena that exhibit power-law behavior.
To see it is a generalization of Shannon entropy, $S_q(\pi)$ can be defined in terms of the deformed $q$-logarithm as follows:
\begin{align}
  S_q(\pi) = \AdaAngleProduct{-\pi^q}{\ln_q \pi}, \quad \ln_q\pi := \begin{cases}
    \ln\pi & \text{if } q=1 \\
    \frac{\pi^{1-q} - 1}{1-q} & \text{if } q \neq 1.
  \end{cases}
  \label{eq:qlog_definition}
\end{align}
$\tsallis{\pi}$ and \qlog have the following properties:
\begin{itemize}
  \item \textbf{Convexity. } $\tsallis{\pi}$ is convex for $q\leq 0$,  concave for $q \geq 0$, and is a linear function for $q=0$.
  \item \textbf{Boundedness. } $ 0 \leq \tsallis{\pi} \leq \ln_q |\mathcal{A}|$ for all $q$, where $|\mathcal{A}|$ denotes the cardinality of the action set.
  \item \textbf{Monotonicity. } $\ln_q\pi$ is monotonically increasing with respect to $\pi$.
  \item \textbf{Pseudo-additivity. } \qlog satisfies:
  \begin{align}
    \ln_q{\pi\mu} = \ln_q\pi + \ln_q \mu + (1-q)\ln_q\pi\ln_q\mu.
    \label{eq:pseudo_add}
  \end{align}
\end{itemize}
Here, convexity ensures the regularization properties analyzed in \citep{geist19-regularized,Li2019-regularizedSparse} hold.
Boundedness allows for derivation of lower- and upper-bounds on the optimal action values \citep{Lee2018-TsallisRAL,Lee2020-generalTsallisRSS}.

Maximizing the general Tsallis entropy $\tsallis{\pi}$ in RL has been considered in \citep{Lee2020-generalTsallisRSS}.
However, taking one step further, employing general Tsallis KL regularization (Tsallis relative entropy) has never seen published results to the best of authors' knowledge.
Nonetheless, we believe its introduction into RL is necessary for rendering MDQN compatible with Tsallis entropy.
Moreover, (generalized) KL divergence is fundamental in the sense that it produces the entropy and the mutual information as special cases \citep{Furuichi2004-fundamentals-qKL}.

With the help of \qlog,  one might hope to correct the mismatch in \eq{\ref{eq:mdqn_qlog}} by modifying the Munchausen term to its $\ln_q$ version,  hence recover implicit Tsallis KL regularization.
However, this point is not straightforward as the result of the pseudo-additivity \eq{\ref{eq:pseudo_add}}: 
\begin{align*}
  \ln_q\!\frac{\pi_{k+1}}{\pi_k} &= \ln_q\pi_{k+1}\cdot\frac{1}{\pi_{k}}\\
  & =  \ln_q\pi_{k+1} + \ln_q \frac{1}{\pi_{k}} + (1-q)\ln_q\pi_{k+1}\ln_q\frac{1}{\pi_{k}}.
\end{align*}
The term $\ln_q\!\frac{\pi_{k+1}}{\pi_k}$ was crucial in forming the implicit KL regularization in \eq{\ref{eq:munchausen_derivation}}.
Unfortunately for \qlog there is a residual $(q-1)\ln_q\pi_{k+1}\ln_q\frac{1}{\pi_{k}}$, which destroys the equivalence required.
Furthermore, the residual involves both the current policy $\pi_{k+1}$ and the base one $\pi_k$, hence one cannot define a new action value same as $Q'_{k+1}$.

Fortunately, by resorting to the $2-q$ duality from the Tsallis statistics literature, we can define Tsallis KL divergence in another way, rendering implicit Tsallis KL regularization possible.

\subsection{Tsallis KL Regularization}\label{sec:implicit_TsallisKL}

\begin{figure}
  \centering
  \includegraphics[width=\linewidth]{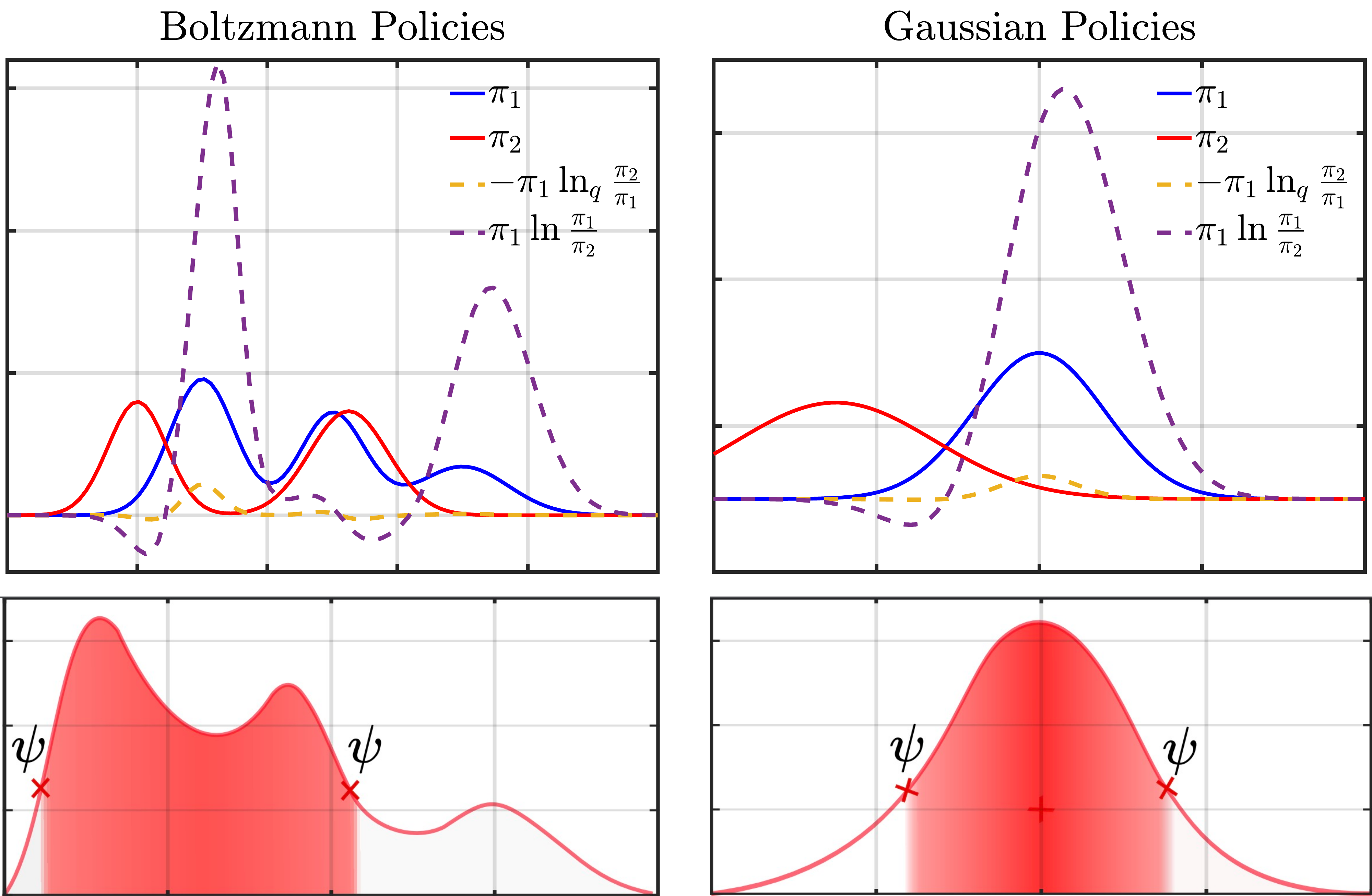}
  \caption{(Top) Illustration of KL and Tsallis KL divergence components $\pi_1\ln\frac{\pi_1}{\pi_2}$ and  $-\pi_1\ln_q{\frac{\pi_2}{\pi_1}}$ between Boltzmann policies and Gaussian policies when $q=2$.
  (Bottom) Illustration of sparsemax acting upon $\pi_1$ by truncating actions that have values lower than $\psi$ defined in \eq{\ref{eq:sparse_normalization}}.
  }
  \label{fig:qkl_examples}
\end{figure}

Before introducing the solution, we need a few definitions.
For any two policies $\pi$ and $\mu$, we define their Tsallis KL divergence as:
\begin{align}
 \qKLany{\pi}{\mu} := - \AdaAngleProduct{\pi}{\ln_q\frac{\mu}{\pi}},
 \label{eq:tsallis_kl}
\end{align}
where it reduces to KL divergence when $q\rightarrow 1$.
Examples of KL and Tsallis KL of the frequently used Boltzmann and Gaussian policies are shown in Figure \ref{fig:qkl_examples}.
 Tsallis KL divergence possesses the following properties similar to that of KL \citep{Furuichi2004-fundamentals-qKL}:
\begin{itemize}
  \item \textbf{Nonnegativity. } $\qKLany{\pi}{\mu} \geq 0$.
  \item \textbf{Convexity. } For $0\leq\lambda\leq 1$ and $q \geq 0$, consider policies $\pi_{i}, \mu_{i}, i=1,2$ and their interpolation $\tilde{\pi}: = \lambda\pi_{1} + (1-\lambda)\pi_{2}$, $\tilde{\mu}:=\lambda\mu_{1} + (1-\lambda)\mu_{2}$,  we have
  \begin{align}
    \qKLany{\tilde{\pi}}{\tilde{\mu}} \leq \lambda\qKLany{\pi_1}{\mu_1} + (1-\lambda)\qKLany{\pi_2}{\mu_2}.
  \end{align}
\end{itemize}
The above properties hold for both KL and Tsallis KL divergences, hence we can expect their regularization have similar characteristics.
For example, it might be possible to replace the theoretical results \citep{kozunoCVI,vieillard2020leverage} with their \qlog version. 

The first step of deriving implicit Tsallis KL regularization is to rewrite $S_q(\pi_{k+1})$ into a compatible form with the inner product $\AdaAngleProduct{\pi_{k+1}}{Q_{k} - \tau\ln_{q^*}\!\pi_{k+1}}$, where $q^*$ is some dual entropic index of $q$: since the original definition $S_q(\pi) = \AdaAngleProduct{-\pi_{k+1}^q}{\ln_q\pi_{k+1}}$ has power expectation which does not easily translate to the regularization scheme in \eq{\ref{eq:regularizedPI}}.
To this end, we adopt the $q^*=2-q$ duality \citep{Naudts2002DeformedLogarithm} which has the equivalence \citep{Suyari2005-LawErrorTsallis}:
\begin{align}
  S_q(\pi_{k+1}) = \AdaAngleProduct{-\pi_{k+1}^q}{\ln_q\pi_{k+1}} = \AdaAngleProduct{-\pi_{k+1}}{\ln_{q^*}\!\pi_{k+1}},
  \label{eq:qlog_equivalence}
\end{align} 
\eq{\ref{eq:qlog_equivalence}} defined a dual $q^*$-logarithm which is consistent with the one used in recent RL literature \citep{Lee2020-generalTsallisRSS}.
The properties of \eq{\ref{eq:qlog_definition}} still hold with by replacing $q=2-q^*$.

The second step concerns expressing Tsallis KL divergence by $q^*$-logarithm without the residual term.
By the $q$-power expectation formulation of Tsallis KL \citep{Ohara2007-geometryDisrtibutionsTsallis} we have 
\begin{align*}
&\qKLany{\pi_{k+1}}{\pi_{k}} := - \AdaAngleProduct{\!\pi_{k+1}}{\ln_q\frac{\pi_{k}}{\pi_{k+1}}\!} \\
&= \AdaAngleProduct{\pi_{k+1}^q}{\ln_q\pi_{k+1} - \ln_q\pi_{k}} 
= \AdaAngleProduct{\pi_{k+1}^{q}}{\frac{ \pi_{k+1}^{1-q} - \pi_{k}^{1-q} }{1 - q}  }.
\end{align*}
We now show the equivalence of $\AdaAngleProduct{\pi_{k+1}^q}{\ln_q\pi_{k+1} - \ln_q\pi_{k}}$ and $\AdaAngleProduct{\pi_{k+1}}{\logqstar{\pi_{k+1}} - \logqstar{\pi_k}}$ by exploiting the $2-q$ duality for the $S_q(\pi)$ \citep[Eq.(12)]{Suyari2005-LawErrorTsallis}:
\begin{align}
  \begin{split}
    &\AdaAngleProduct{\pi_{k+1}}{\logqstar{\pi_{k+1}} \!-\! \logqstar{\pi_k}} \!=\! \AdaAngleProduct{\!\! \pi_{k+1}}{ \frac{\pi_{k+1}^{q^*-1} - 1}{q^* - 1} \!-\! \frac{\pi_{k}^{q^*-1} - 1}{q^* - 1} \!}\\
    &= \AdaAngleProduct{\! \pi_{k+1}}{ \frac{\pi_{k+1}^{q^*-1} - \pi_{k}^{q^*-1}}{q^* - 1}  \!} =  \AdaAngleProduct{\! -\pi_{k+1}}{ \frac{  \AdaBracket{ \frac{\pi_{k}}{\pi_{k+1}} }^{q^*-1} \!- 1 }{\pi_{k+1}^{1-q^{*}}(q^* - 1) }  \!} \\
    &= \AdaAngleProduct{\! -\pi_{k+1}^{2-q^*}}{ \frac{  \AdaBracket{ \frac{\pi_{k}}{\pi_{k+1}} }^{q^*-1} \!- 1 }{q^* - 1 }  \!} 
    \!=\! \AdaAngleProduct{\!\! -\pi_{k+1}^{q}}{ \frac{  \AdaBracket{ \frac{\pi_{k}}{\pi_{k+1}} }^{1-q} \!- 1 }{1-q }  \!} \\
    &= \qKLindex{k+1}{k},
  \end{split}
\end{align}
Given the above equation, we can now reformulate the Tsallis version of MDQN as:
\begin{align}
  \begin{split}
    &Q_{k+1} = r + {\color{red} \alpha\tau \logqstar{\pi_{k+1}} }  + \gamma P \AdaAngleProduct{\pi_{k+1}}{Q_k - {\color{blue} \tau\logqstar{\pi_{k+1}} } },\\
    &\Leftrightarrow Q_{k+1} - \alpha\tau\logqstar{\pi_{k+1}} = \AdaAngleProduct{\pi_{k+1}}{Q_k - \alpha\tau\logqstar{\pi_k}} - \\ 
    & \qquad \AdaAngleProduct{\pi_{k+1}}{ \alpha\tau(\logqstar{\pi_{k+1}} -  \logqstar{\pi_{k}}) - (1-\alpha)\tau\logqstar{\pi_{k+1}} },\\
    & \Leftrightarrow Q''_{k+1} = \\
    & r \!+\! \gamma P\! \AdaAngleProduct{\pi_{k+1}}{Q''_k} \!-\! \alpha\tau\qKLindex{k+1}{k} \!+\!  (1-\alpha)\tau S_q(\pi_{k+1}),
  \end{split}
  \label{eq:general_mdqn}
\end{align}
where $Q''_{k+1}\!:=\! Q_{k+1} - \alpha\tau\logqstar{\pi_{k+1}}$ is the new action value function.
Hence it is clear that \eq{\ref{eq:general_mdqn}} generalizes MDQN in \eq{\ref{eq:munchausen_derivation}} by letting $q=1$.

\subsection{Tsallis Policies}\label{sec:tsallis_policy}

Similar with the exponential policy form induced by Shannon entropy, $q^*$-logarithm induces $q^*$-exponential $\expqstar{}$ as its inverse function:
\begin{align}
  \expqstar{(Q)} = \begin{cases}
    \exp(Q), & \text{if }  q=1 \\
    \AdaRectBracket{1 + (q^* - 1)Q}^{\frac{1}{q^*-1}}_{+}, & \text{if } q\neq 1.
  \end{cases}
\end{align}
$q^*$-exponential has been exploited in \citep{Lee2020-generalTsallisRSS} to yield the maximum Tsallis entropy policies as $\pi^{*}_{q^*} \!=\! \expqstar{(Q/q^* - \psi_{q^*}(Q)/q^*)}$, where $\psi_{q^*}$ is the normalization term depending on specific choices of the dual entropic index $q^*$.
When $q^*\neq 1,2,\infty$, there might be no closed-form expression for $\psi_{q^*}$ and hence the policy $\pi^*_{q^*}$.


Motivated by the property of KL regularized policy
$
  \pi_{k+1} \!\propto\! \pi_k\exp(Q_k) \!\propto\! \pi_{k-1}\exp(Q_k + Q_{k-1}) \!\propto\! \dots \!\propto \!\exp(\sum_{j=1}^{k}Q_j),
$
where we drop the regularization coefficient $\tau$ for convenience,
it is natural to expect the Tsallis KL regularized policies possess a similar averaging effect.
This indeeds holds for the \qstarexp since we have \citep{Yamano2004-properties-qlogexp}:
\begin{align}
  \begin{split}
  \expqstar{\AdaBracket{ \sum_{j=1}^{k}Q_j} }^{q^*-1} & \!\!\!\!\!\!= 
   \AdaBracket{ \expqstar{Q_1}\dots\expqstar{Q_k} }^{q^* - 1} \!\! \\
   & + \sum_{j=2}^{k}(q^* - 1)^j \!\!\!\sum_{i_{1} < \dots < i_j}^k \! Q_{i_1} \dots Q_{i_j},
\end{split}
\label{eq:pseudo_average}
\end{align}
When $q^* \!=\! 2$, \eq{\ref{eq:pseudo_average}} reduces to:
\begin{align*}
  \expqstar{\AdaBracket{ \sum_{j=1}^{k}Q_j} \!\!} \!=
   \expqstar{Q_1}\dots\expqstar{Q_k}  \!  + \!\!\!\!\!\! \sum_{\substack{j=2 \\ i_{1} < \dots < i_j}}^{k} \!\!\!\! Q_{i_1} \dots Q_{i_j},
\end{align*}
in which the residual term $\sum_{\substack{i_{1} < \dots < i_j}}^{k} Q_{i_1} \dots Q_{i_j}$ can be regarded as a special case of adaptively weighting the contribution of each $Q_j$ function rather than the uniform weighting.
One can also define the $q^*$-product $\times_{q^*}$ \citep[Eq.(29)]{Suyari2005-LawErrorTsallis} so that $\expqstar{Q_1} \times_{q^*}\expqstar{Q_2} = \expqstar{(Q_1+Q_2)}$.

\subsection{Practical Implementation}\label{sec:implementation}

\begin{algorithm}[t]
  \caption{Tsallis Entropy Munchausen DQN}\label{algorithm}
  \SetKwInput{KwInput}{Input}
  \SetKwInput{KwInit}{Initialize}
  \KwInput{total number of steps $T$, update period $I$, \\ interaction period $C$, epsilon-greedy threshold $\epsilon$, \\
  entropic index $q^*$, entropy coefficients $\alpha,\tau$;
  }
  \KwInit{ Network weights $\bar{\theta} = \theta$; \\ 
  Replay buffer $B = \{\}$;
  }
  \For{$t = 1, 2, \dots, T$ }
  {
      Collect tuple $(s_t, a_t, r_t, s_{t+1})$ in $\mathcal{B}$ with $\pi_{q^*, \epsilon}$\\
      \If{ mod$(t, C) = 0$}
      {   
          sample a minibatch $\mathcal{B}_t \subset \mathcal{B}$ \\
          compute $Q_{\bar{\theta}}, \pi_{\bar{\theta}}$ using target network $\bar{\theta}$ \\
          compute $\logqstar{\pi_{\bar{\theta}}}=\frac{\pi_{\bar{\theta}}^{q^*-1}-1}{q^*-1}$\\
          update $\theta$ with one step of SGD on the loss $\mathcal{L}$ \\
      }   
      \If{mod$(t, I) = 0$}
      {
          update target network $\bar{\theta} \leftarrow \theta$\\
      }
  }
\end{algorithm}

Given \eq{\ref{eq:general_mdqn}}, we detail our proposed algorithm: Tsallis Entropy Munchausen DQN (TEMDQN) by modifying the TD loss of MDQN.
Specifically, we replace every appearance of $\ln\pi_{k+1}$ in \eq{\ref{eq:mdqn_recursion}} by $q^*$-logarithm $\logqstar{\pi_{k+1}}$, which leads to the following loss function:
  \[
    \mathcal{L}(\theta) \!:=\! 
    \hat{\mathbb{E}}_{\mathcal{B}}\!\AdaRectBracket{\! \AdaBracket{r \!+\! \alpha\tau\logqstar{\pi_{\bar{\theta}}} \!+\! \gamma\AdaAngleProduct{\pi_{\bar{\theta}}}{Q_{\bar{\theta}} \!-\! \tau\logqstar{\pi_{\bar{\theta}}} }   \!-\! Q_{\theta} }^2 \!},
    \]
where $\hat{\mathbb{E}}_{\mathcal{B}}$ denotes empirical expectation w.r.t. to the replay buffer $\mathcal{B}$.
The policy $\pi_{\bar{\theta}}$ is computed from $Q_{\bar{\theta}}$ as introduced in Section \ref{sec:entropy}.
Note that in this paper we focus on the case $q^*=2$ which corresponds to the Tsallis sparse policy since it enjoys closed-form policy expression.
However, it is straightforward to extend TEMDQN to other entropic indices by employing approximate policy following \citep{chen2018-TsallisApproximate,Lee2020-generalTsallisRSS}.

The pseudo-code of TEMDQN is listed in Alg. \ref{algorithm}.
At every step $t$, the environment is explored using the policy $\pi_{q^*, \epsilon}$ to collect a tuple of experience into the buffer $\mathcal{B}$, where $\epsilon$ denotes a small value used for $\epsilon$-greedy exploration.
Every $C$ steps, the network is updated by running stochastic gradient descent on a randomly sampled batch $\mathcal{B}_t$ on the loss function $\mathcal{L}$.
The target network is updated every $I$ steps.
Implementation details such as hyperparameters are provided in Appendix \ref{apdx:experiment_setting}.

\section{Experiments}\label{sec:experiment}

We aim to validate the proposed method  on both simple and challenging benchmark problems.
Specifically, we show that MDQN can perform significantly poor with Tsallis sparse policy due to the logarithm mismatch, while TEMDQN still achieves superior performance.

\begin{figure}
  \centering
  \includegraphics[width=\linewidth,trim=4 4 4 4,clip]{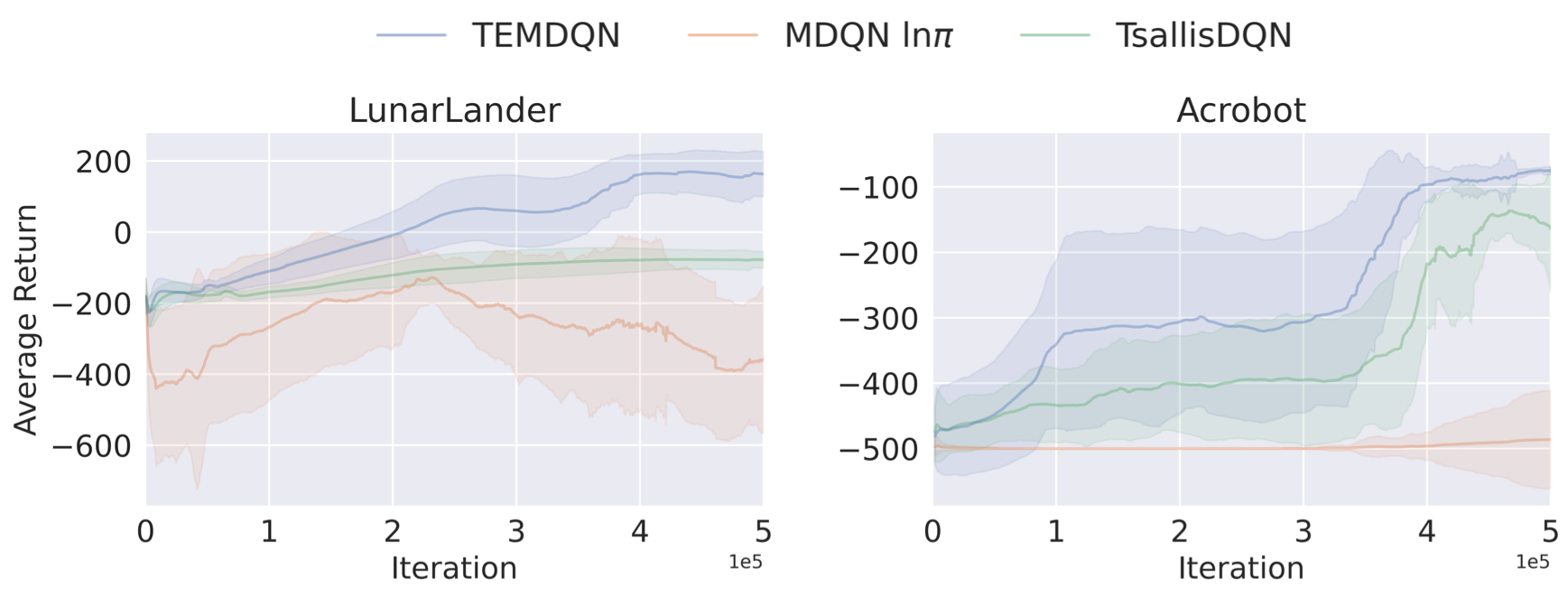}
  \caption{Comparison between TEMDQN, log-sparsemax MDQN and TsallisDQN on \texttt{LunarLander-v2} and \texttt{Acrobot-v1} with the policy being Tsallis sparsemax. 
  MDQN $\logqstar{\pi}$ denotes the TEMDQN augmentation.
  All algorithms are averaged over 50 independent trials to plot the mean and $\pm 1$ standard deviation.
  }
  \label{fig:gym}
\end{figure}

Figure \ref{fig:gym} shows the comparison between TEMDQN, log-sparsemax MDQN and TsallisDQN  on the classic control problems from the OpenAI Gym \citep{brockman2016openai}, with implementation details provided in Appendix \ref{apdx:gym}.
It serves as the basic validation for our claim, from which we see log-sparsemax MDQN failed to solve the tasks, while TEMDQN improved upon TsallisDQN and successfully converged to the optimum.

We also compared TEMDQN and log-sparsemax MDQN in more challenging environments such as the MiniGrid \citep{gym_minigrid}, MinAtar games \citep{young19minatar}.
We are also interested in examining whether TEMDQN can perform favorably with distributional methods by running it on the Atari games \citep{bellemare13-arcade-jair}.
For reliable and reproducible evaluation, we implement all algorithms based on the Stable-Baselines3 library \citep{stable-baselines3}.

\subsection{MiniGrid Environments}\label{sec:minigrid}

We aim to further consolidate the effectiveness of TEMDQN on more challenging tasks such as the MiniGrid environments featuring sparse reward, partial observability, the need for efficient exploration and visual input \citep{gym_minigrid}. 
To better illustrate the effectiveness, we choose \texttt{Empty-16$\times$16} and \texttt{Dynamic-Obstacles-16$\times$16} which are the biggest environments of their kinds.
The agent in \texttt{Empty} starts from a fixed position and attempts to reach the goal given partial observation and negative reward alone the way.
In \texttt{Dynamic-Obstacles}, the environment is equipped with randomly moving obstacles and running into them incurs a large penalty.
Implementation details are provided in Appendix \ref{apdx:minigrid}.

\begin{figure}[t]
  \centering
  \includegraphics[width=\linewidth,trim=4 4 4 4,clip]{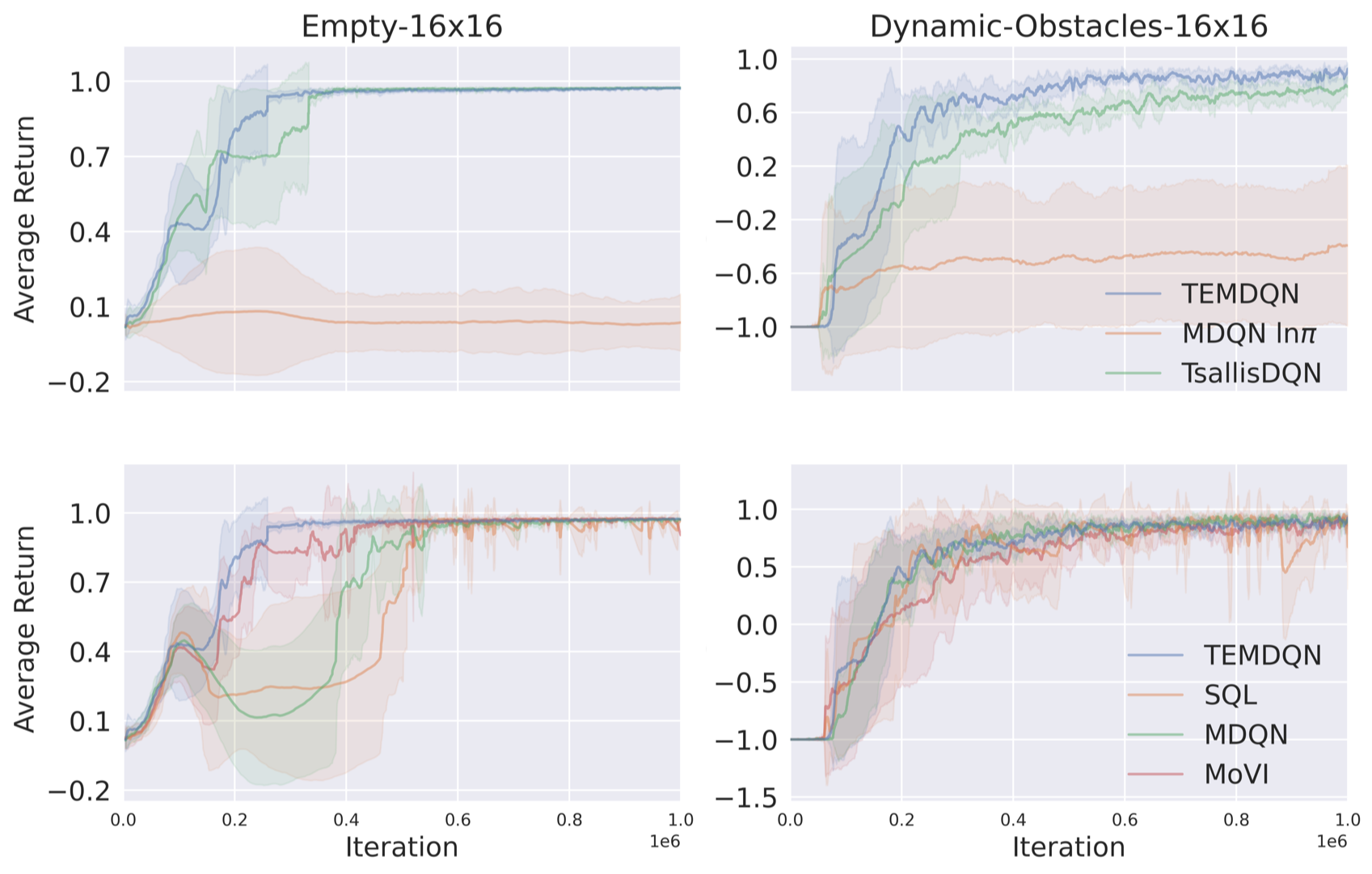}
  \caption{
    Comparison on the selected MiniGrid environments. 
    (Top) TEMDQN, log-sparsemax MDQN and TsallisDQN.
    (Bottom) TEMDQN, SQL, original MDQN and MoVI.
  }
  \label{fig:minigrid_tsallis}
\end{figure}

\textbf{TsallisDQN Comparison. }
We first compare TEMDQN against TsallisDQN and log-sparsemax MDQN. 
For statistical significance, all algorithms are averaged over 10 independent trials to plot mean and $\pm 1$ standard deviation in Figure \ref{fig:minigrid_tsallis}.
Consistent with our analysis, it is visible that log-sparsemax MDQN failed to learn meaningful behaviors in \texttt{Empty-16$\times$16}, leading to a flat learning curve, while TEMDQN benefited greatly from the implicitly Tsallis KL regularization and quickly converged to the optimum.
TsallisDQN also managed to converge to the optimum, but at a slower rate than TEMDQN.

In \texttt{Dynamic-Obstacles-16$\times$16}, the $\ln\pi$ augmentation again failed to lead MDQN to the goal due to the logarithm mismatch.
On the other hand, TsallisDQN converged to a sub-optimum as a result of dynamic obstacles, which resulted in slower convergence and lower final score than TEMDQN.



\begin{figure}[t]
  \centering
  \includegraphics[width=1.0\linewidth,trim=4 4 4 4,clip]{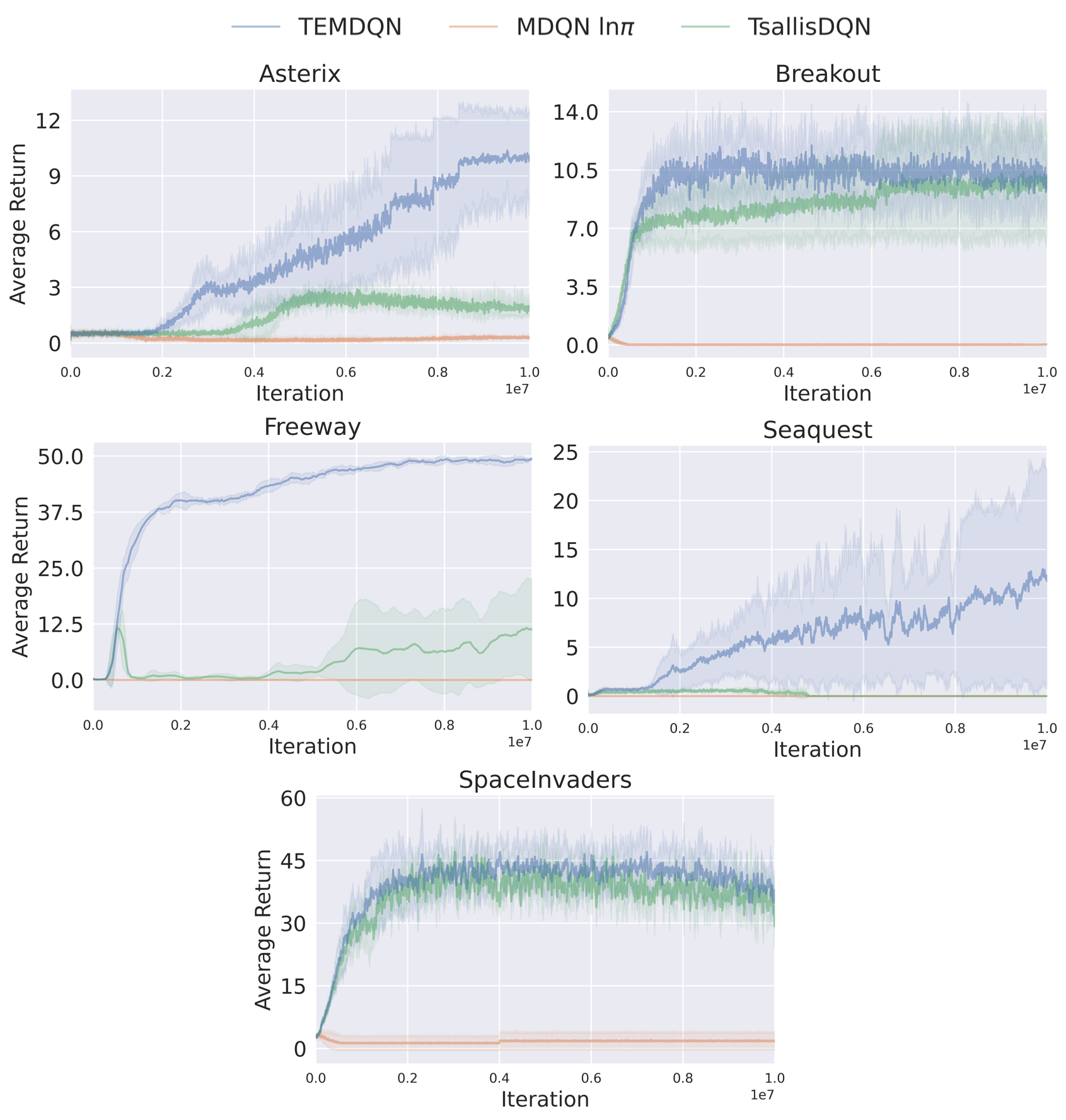}
  \caption{
    Comparison of TEMDQN against TsallisDQN and log-sparsemax MDQN on MinAtar environments. 
  }
  \label{fig:minatar}
\end{figure}

\begin{figure*}
  \centering
  \includegraphics[width=0.99\textwidth,trim=4 4 4 4,clip]{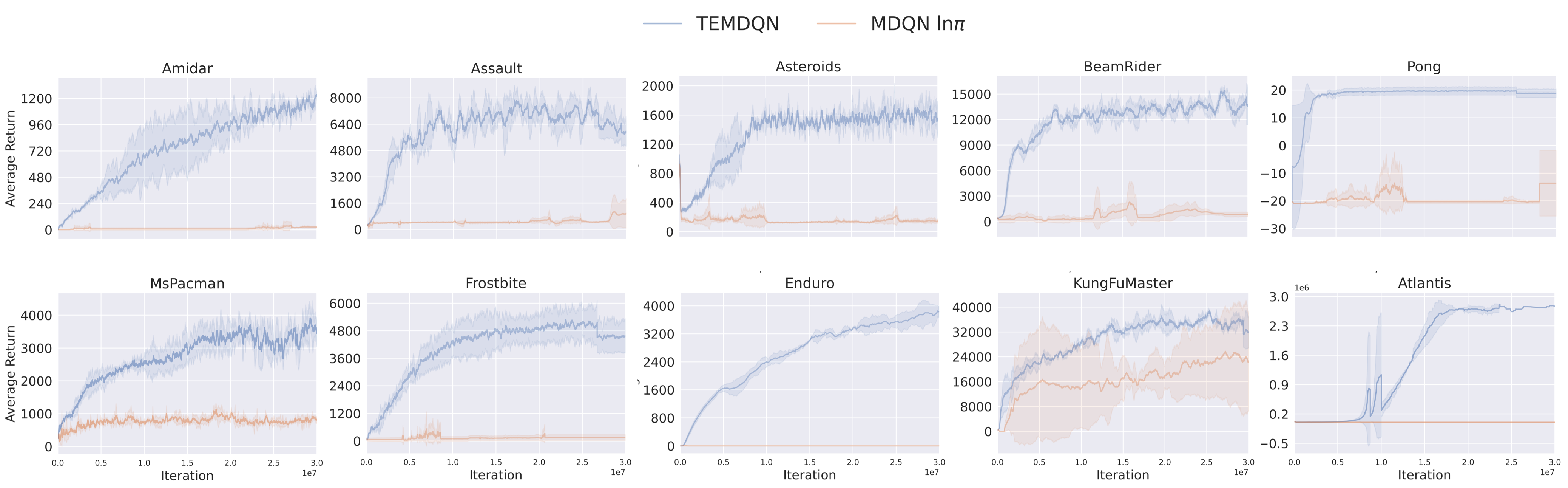}
  \caption{Comparison between distributional TEMDQN and log-sparsemax MDQN on Atari games.
  }
  \label{fig:distributional}
\end{figure*}

\textbf{Averaging. }
One of the most important features of KL regularization is the averaging effect when facing errors.
It is hence interesting to inspect whether TEMDQN benefits from the \emph{pseudo-averaging} effect introduced in \eq{\ref{eq:pseudo_average}}.
In this subsection we compare TEMDQN against existing state-of-the-art algorithms featuring such averaging effect.
Specifically, we compare against:
\begin{itemize}
  \item \text{Speedy Q-learning (SQL)} \citep{Azar2011-speedy} which reduces the greediness of Q-learning by using two value estimates. 
  It is shown in \citep{vieillard2020leverage} that SQL policy can be expressed as $\argmax_{\pi}\AdaAngleProduct{\pi}{\frac{1}{k} \sumJK{1}Q_j - \frac{\tau}{k} \ln\pi}$.
  \item \text{Momentum Value Iteration (MoVI)} \citep{Vieillard2020Momentum}  which approximates KL policy without the help of Shannon entropy.
  MoVI policy is expressed as $\argmax_{\pi}\AdaAngleProduct{\pi}{\frac{1}{k}\sumJK{1}Q_j}$.
  \item \text{MDQN. } We compare with the original MDQN implemented based on soft Q-learning, which achieved significant improvement.
\end{itemize}

The comparison is shown in the bottom of Figure \ref{fig:minigrid_tsallis}.
In \texttt{Empty}, it is surprising to see that TEMDQN reached to the optimum faster than all other algorithms.
The slower convergence of MDQN and SQL might be due to the Shannon entropy augmentation assigned probabilities to actions not helpful to reaching the goal.
We compare the the values of $\ln\pi$ and $\logqstar{\pi}$ for the original MDQN and TEMDQN in Figure \ref{fig:minigrid_logpi} in Appendix \ref{apdx:minigrid}.

In \texttt{Dynamic-Obstacles}, MDQN and SQL converged faster thanks to both the averaging as well as diversity, which is confirmed by the slightly inferior MoVI deterministic policy.
By contrast, TEMDQN achieved a good balance leading to fast convergence in both domains.

\subsection{MinAtar Games}\label{sec:minatar}

Atari games are one of the standard challenging benchmark problems for examining modern RL algorithms.
In this subsection we aim to show TEMDQN can also effectively learn in such problems.
We adopt the MinAtar environments \citep{young19minatar} which consist of five Atari games optimized comprehensively e.g. frame-skipping, observation space, etc.
TEMDQN is compared against TsallisDQN and log-sparsemax MDQN for $10^7$ frames without frame skip.
Experimental details are provided in Appendix \ref{apdx:experiment_setting}.
Results are averaged over 5 seeds.

From Figure \ref{fig:minatar} it is visible that the trend in previous experiments continue to hold: log-sparsemax MDQN led to flat learning curves for all games due to the mismatch.
On the other hand, TEMDQN was capable of improving upon TsallisDQN by a large margin on hard exploration games \texttt{Asterix, Freeway, Seaquest}. 
The underperformant behavior of TsallisDQN on those environments is expected since the sparsemax policy can result in insufficient exploration which is critical to Atari games \citep{Lee2020-generalTsallisRSS}.
Though the improvement for the other two games was not so significant, it is relieving that TEMDQN never perform worse than the TsallisDQN, rendering it a safe substitute for TsallisDQN.

\subsection{Distributional Munchausen}\label{sec:distributional}

We implemented TEMDQN and log-sparsemax MDQN on top of Quantile Regression DQN (QR-DQN) \citep{Dabney2018-QRDQN} and compare them on a subset of 10 full-fledged Atari games \citep{bellemare13-arcade-jair}.
Instead of learning the conditional expectation $Q$ function, distributional RL attempts to learn the distribution itself:
\begin{align}
    \begin{split}
    &Q_{\pi}(s,a) = \mathbb{E}\AdaRectBracket{Z_{\pi}(s,a)},\\
    &\Bellman{}Z(s,a) = r(s,a) + \gamma Z(s',a').
\end{split}
\end{align}
Specifically, QR-DQN aims to approximate $Z$ by regressing state-action pairs $(s,a)$ to a uniform probability distribution supported by $N$ fixed quantiles $\varphi_{i}: Z_{\varphi}(s,a) := \frac{1}{N}\sum_{j=1}^{N}\delta_{\varphi_j(s,a)}$, where $\delta_{\{\cdot\}}$ is a Delta function.
The implementations of distributional TEMDQN and log-sparsemax MDQN replace every appearance of $Q$ with $Z$ in Eqs. (\ref{eq:mdqn_recursion}), (\ref{eq:general_mdqn}).
Implementation details are provided in Appendix \ref{apdx:atari}.

We allow $3 \times 10^7$ environment steps for training and the results are averaged over three seeds and shown in Figure \ref{fig:distributional}.
It is visible that log-sparsemax MDQN has the same trend as in prior experiments of yielding nearly flat learning curves, which is outperformed by TEMDQN by a large margin.

\section{Discussion and Conclusion}\label{sec:conclusion}


In this paper we showed that Munchausen RL with log-sparsemax-policy augmentation exhibited significantly poor performance for almost every problem considered, which contradicted the general claim of M-RL that arbitrary stochastic policy could be used.
We showed that the poor performance was due to the mismatch between the conventional logarithm and the non-logarithmic (power) nature of Tsallis entropy.
Inspired by the Tsallis statistics literature, we reformulated Tsallis entropy and rewrote the M-RL update rule by $q$-logarithm.
The resultant algorithm enjoyed implicit Tsallis KL divergence regularization which has neven been considered in RL literature.
Experiments on various benchmark problems validated the effectiveness of the proposed method.


One interesting future direction is to examine various ways of implicit Tsallis KL regularization by following a similar idea to existing literature such as MoVI, SQL, and compare their empirical performance.





\clearpage

\bibliography{../../library}

\clearpage

\appendix

\section{Experimental Setting}\label{apdx:experiment_setting}

Due to their respective characteristics, we exploited three different architectures for running experiments on gym environments, MiniGrid and MinAtar games and Atari games, detailed in Sections \ref{apdx:gym}, \ref{apdx:minigrid}, \ref{apdx:minatar} and \ref{apdx:atari}, respectively.

\subsection{Gym Classic Control }\label{apdx:gym}

The Gym classic control environments are an ideal testbed of evaluating the correctness of the proposed method.
We select the environments \texttt{LunarLander-v2} and \texttt{Acrobot-v1} for comparison which are reasonably complicated and can be solved by fully connected networks.
TEMDQN is compared against TsallisDQN (sparsemaxDQN) \citep{Lee2018-TsallisRAL} and log-sparsemax MDQN, where $\pi$ is the Tsallis sparsemax policy induced by $S_2(\pi)$.
Totally $5\times 10^5$ steps are allowed for solving both tasks, and we perform 50 independent trials to plot the mean and $\pm 1$ standard deviation in Figure \ref{fig:gym}.


The hyperparameters used for the \texttt{LunarLander-v2} and \texttt{Acrobot-v1} are detailed in Table \ref{tb:gym}.
The epsilon threshold is fixed at $0.01$ from the beginning of learning. 
Since the sparsemax policy assigns $0$ to some actions, the $\ln\pi$ augmentation might be undefined.
To this end,  we add a small value $\Delta$ to the policy to prevent ill-defined $\ln\pi$.
FC$\,n$ refers to the fully connected layer with $n$ activation units.
To find a performant set of entropy coefficient $\tau$ and Munchausen augmentation coefficient $\alpha$, we performed a grid search over the sets $\tau = \{10^{-4}, 10^{-3}, 10^{-2}, 10^{-1}, 10^0, 10^1, 10^2 \}$ and $\alpha = \{10^{-4},  10^{-3}, 0.01, 0.1, 0.5, 0.9\}$.

\begin{table}[h!]
    \centering
    \caption{Parameters used for TEMDQN, log-sparsemax MDQN and TsallisDQN on \texttt{LunarLander-v2} and \texttt{Acrobot-v1}. }
    \label{tb:gym}
    \begin{tabular}{llr}
        \cline{1-2}
        Parameter    & Value \\
        \hline
        $T$ (total steps)     &  $5\times 10^5$     \\
        $C$ (interaction period) &      4      \\
        $|\mathcal{B}|$ (buffer size)      & $5 \times 10^4$     \\
        $\mathcal{B}_t$ (batch size)  &  128 \\
        $\gamma$ (discount rate)       & 0.99      \\
        $I$  (update period)      &  $2500$ \\
        $\epsilon$ (epsilon greedy threshold) & 0.01 \\
        $\Delta$ (ill-defined $\ln\pi$ prevention number) & $10^{-8}$\\
        $\tau$ (Tsallis entropy coefficient) &  10 \\
        $\alpha$ (Munchausen coefficient) & 0.0025 \\
        Q-network architecture & FC512 - FC512 \\
        activation units & ReLU \\
        optimizer & Adam \\
        optimizer learning rate & $10^{-3}$ \\
        \hline
    \end{tabular}
\end{table}

\subsection{MiniGrid Environments}\label{apdx:minigrid}

\begin{figure}[t]
  \centering
  \includegraphics[width=\linewidth,trim=4 4 4 4,clip]{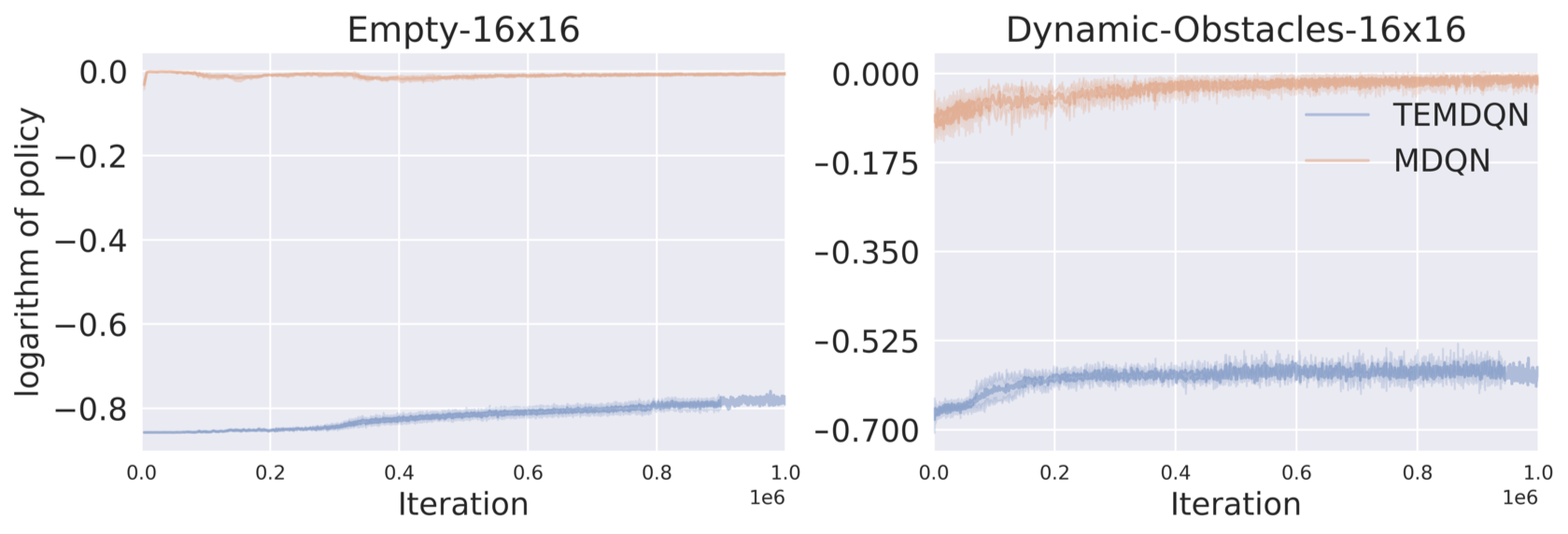}
  \caption{
    Comparison between $\ln\pi$ and $\logqstar{\pi}$ on MiniGrid environments. TEMDQN log-policy tends to be of larger magnitude than MDQN.
  }
  \label{fig:minigrid_logpi}
\end{figure}

The algorithms performed on the MiniGrid environments took images of the gridworld as input and requires sophisticated network architecture for processing.
We follow the architecture in \citep{Yannis2021-AGAC} and show the details in Table \ref{tb:minigrid}.
To find a performant set of parameters $\tau$ and $\alpha$,
we performed a grid search over $\tau = \{2.5\times 10^{-4}, 2.5\times 10^{-3}, 0.025, 0.25, 2.5\}$ and $\alpha = \{10^{-4}, 10^{-3}, 10^{-2}, 0.1, 0.5, 0.9\}$.
We compared the the values of $\ln\pi$ and $\logqstar{\pi}$ for the original MDQN and TEMDQN in Figure \ref{fig:minigrid_logpi}.
It is clear that the magnitude of $\logqstar{\pi}$ tends to be large while MDQN prefers small log-policy, which could be interpreted as TEMDQN prefers larger Tsallis KL regularization effect.

\begin{table}[hb!]
  \centering
  \caption{Parameters used for the MiniGrid environments. }
  \label{tb:minigrid}
  \begin{tabular}{llr}
      \cline{1-2}
      Parameter    & Value \\
      \hline
      $T$ (total steps)     &  $1 \times 10^6$     \\
      $C$ (interaction period) &      4     \\
      $|\mathcal{B}|$ (buffer size)      & $1 \times 10^6$     \\
      $\mathcal{B}_t$ (batch size)  &  256 \\
      $\gamma$ (discount rate)       & 0.99      \\
      $I$  (update period)      & $2048$  \\
      $\epsilon$ (epsilon greedy threshold) & $1.0 \rightarrow 0.01 |_{10\%}$ \\
      $\Delta$ (ill-defined $\ln\pi$ prevention number) & $10^{-8}$\\
      $\tau$ (Tsallis entropy coefficient) &  0.25 \\
      $\alpha$ (Munchausen coefficient) & 0.0001 \\
      $\beta$ (SQL/MoVI mixture steps) & $2.5\times 10^4$ \\
      Q-network architecture & \\ 
      \multicolumn{2}{c}{\quad FC512 - $\text{Conv}^{2}_{3,3}32$ - $\text{Conv}^{2}_{3,3}32$ - $\text{Conv}^{2}_{3,3}32$ - FC}  \\
      activation units & ELU \\
      optimizer & Adam \\
      optimizer learning rate & $3 \times 10^{-4}$\\
      \hline
  \end{tabular}
\end{table}

\subsection{MinAtar Games}\label{apdx:minatar}

For MinAtar games we employed the configuration recommended by \cite{young19minatar}.
The hyperparameters are listed in Table \ref{tb:minatar}.

Since MinAtar games optimize different aspects of Atari games, no frame skipping is necessary and hence the for every frame we update the network. 
The epsilon greedy threshold $\epsilon: 1.0 \rightarrow 0.05 |_{10\%}$ denotes that $\epsilon$ is initialized as 1.0 and gradually decays to $0.05$ through the first 10\% of learning.
The network architecture consists of only one convolutional layer where $\text{Conv}^{d}_{a,b}c$ denotes a convolutional layer with $c$ filters of size $a\times b$ and stride $d$.
To find a performant set of parameters $\tau$ and $\alpha$,
we performed a grid search over $\tau = \{2.5\times 10^{-4}, 2.5\times 10^{-3}, 0.025, 0.25, 2.5\}$ and $\alpha = \{10^{-4}, 10^{-3}, 10^{-2}, 0.1, 0.5, 0.9\}$.

\begin{table}
    \centering
    \caption{Parameters used for TEMDQN, log-sparsemax MDQN and TsallisDQN on all MinAtar games. }
    \label{tb:minatar}
    \begin{tabular}{llr}
        \cline{1-2}
        Parameter    & Value \\
        \hline
        $T$ (total steps)     &  $1 \times 10^7$     \\
        $C$ (interaction period) &      1      \\
        $|\mathcal{B}|$ (buffer size)      & $1 \times 10^5$     \\
        $\mathcal{B}_t$ (batch size)  &  32 \\
        $\gamma$ (discount rate)       & 0.99      \\
        $I$  (update period)      & $1000$  \\
        $\epsilon$ (epsilon greedy threshold) & $1.0 \rightarrow 0.05 |_{10\%}$ \\
        $\Delta$ (Munchausen number) & $10^{-8}$\\
        $\tau$ (Tsallis entropy coefficient) &  0.025 \\
        $\alpha$ (Munchausen coefficient)  & 0.01 \\
        Q-network architecture & $\text{Conv}^{1}_{3,3}16$ - FC128 \\
        activation units & ReLU \\
        optimizer & RMSProp \\
        optimizer learning rate & $2.5\times 10^{-4}$\\
        (RMSProp) squared momentum & 0.95 \\
        (RMSProp) minimum momentum & 0.01 \\
        \hline
    \end{tabular}
\end{table}

\subsection{Atari Games}\label{apdx:atari}

\begin{table}[hbt!]
  \centering
  \caption{Parameters used for distributional TEMDQN and log-sparsemax MDQN on Atari games. }
  \label{tb:atari}
  \begin{tabular}{llr}
      \cline{1-2}
      Parameter    & Value \\
      \hline
      $T$ (total steps)     &  $3 \times 10^7$     \\
      $C$ (interaction period) &      4     \\
      $|\mathcal{B}|$ (buffer size)      & $1 \times 10^6$     \\
      $\mathcal{B}_t$ (batch size)  &  32 \\
      $\gamma$ (discount rate)       & 0.99      \\
      $I$  (update period)      & $8000$  \\
      $\epsilon$ (epsilon greedy threshold) & $1.0 \rightarrow 0.01 |_{10\%}$ \\
      $\Delta$ (Munchausen number) & $10^{-8}$\\
      $\tau$ (Tsallis entropy coefficient) &  0.025 \\
      $\alpha$ (Munchausen coefficient) & 0.001 \\
      Q-network architecture & \\ 
      \multicolumn{2}{c}{\quad $\text{Conv}^{4}_{8,8}32$ - $\text{Conv}^{2}_{4,4}64$ - $\text{Conv}^{1}_{3,3}64$ - FC512 - FC}  \\
      activation units & ReLU \\
      optimizer & Adam \\
      optimizer learning rate & $10^{-4}$\\
      \hline
  \end{tabular}
\end{table}


We compared distributional TEMDQN and log-sparsemax MDQN on Atari games which are more challenging and high dimensional than the optimized MinAtar games.
We leveraged the optimized Stable-Baselines3 architecture \citep{stable-baselines3} for best performance. 
The details can be seen from Table \ref{tb:atari}.
The Q-network uses 3 convolutional layers.
The epsilon greedy threshold is initialized at 1.0 and gradually decays to 0.01 at the end of first 10\% of learning.
To find a performant set of parameters $\tau$ and $\alpha$,
we performed a grid search over $\tau = \{2.5\times 10^{-4}, 2.5\times 10^{-3}, 0.025, 0.25, 2.5\}$ and $\alpha = \{10^{-3}, 10^{-2}, 0.1, 0.9\}$.

\end{document}